\documentclass[12pt]{article}
\usepackage{times}
\usepackage[utf8]{inputenc}
\usepackage[margin=1in]{geometry}
\usepackage{titlesec}
\usepackage{multirow}
\usepackage{amsmath}
\usepackage{mathtools}
\usepackage{hyperref}
\usepackage{color}
\usepackage{color,soul}
\usepackage{titlesec}
\usepackage{graphics}
\usepackage{graphicx}
\usepackage{amssymb}
\usepackage{array}
\usepackage{multirow}
\usepackage{float}
\usepackage{caption}
\usepackage{subcaption}
\usepackage{float}
\usepackage[table]{xcolor}
\usepackage{tabu}
\usepackage{booktabs}
\usepackage{rotating}
\usepackage{placeins}
\usepackage{titling}
\usepackage{etoolbox}
\patchcmd{\thebibliography}
 {\advance\leftmargin\labelsep}
 {\leftmargin=0pt\itemindent=\labelwidth\advance\itemindent\labelsep}
 {}{}
\usepackage{url}
\usepackage{authblk}
\usepackage{amsmath,array,ragged2e,booktabs}
\usepackage{adjustbox,lipsum}
\usepackage{chngpage}
\usepackage{epstopdf}
\usepackage{algorithm}
\usepackage{algcompatible}
\usepackage[noend]{algpseudocode}
\usepackage[numbers]{natbib}
\usepackage[font=footnotesize,labelfont=bf]{caption}
\usepackage{fancyhdr}
\usepackage{amsmath}
\usepackage{bbm}
\usepackage{breqn}
\usepackage{natbib}
\usepackage{multicol}
\usepackage{resizegather}
\usepackage{cuted}
\usepackage[nodisplayskipstretch]{setspace}
\usepackage{capt-of}
\graphicspath{{images/}}
\usepackage{amsmath}
\usepackage{amssymb}
\usepackage{txfonts}
\usepackage{algorithm,algpseudocode}

\newcommand{\mc}{\mathcal}
\newcommand{\set}[1]{\left\{ #1 \right\}}
\newcommand{\reals}{\mathbb{R}}
\DeclareMathOperator*{\argmin}{arg\,min}

\Large\title{\textbf{A Machine Learning Approach to \\ 
		Shipping Box Design}}
\date{ }
\let\OLDthebibliography\thebibliography
\renewcommand\thebibliography[1]{
 \OLDthebibliography{#1}
 \setlength{\parskip}{0pt}
 \setlength{\itemsep}{0pt plus 0.3ex}
}

\makeatletter
\def\BState{\State\hskip-\ALG@thistlm}
\makeatother

\newcolumntype{C}{>{\Centering\arraybackslash}m{0.1\linewidth}}

\titleformat*{\section}{\normalsize\bfseries}
\titleformat*{\subsection}{\normalsize\bfseries}
\titleformat*{\subsubsection}{\small\bfseries}

\begin{document}

\setlength{\abovedisplayskip}{4pt}		\setlength{\belowdisplayskip}{4pt}
\vspace{-.15in}
	\author[Author 1 et al.]
	{\rmfamily\normalsize\textbf{Guang Yang}\\\vspace{-.15in}
	\rmfamily\normalsize {Jet.com/Walmart Labs, guang@jet.com}\\\vspace{.15in}
	\rmfamily\normalsize\textbf{Cun (Matthew) Mu}\\
	\rmfamily\normalsize {Jet.com/Walmart Labs, matthew.mu@jet.com}\\
}

	\maketitle
\thispagestyle{fancy}
	\vspace{-.5in}

\begin{center}
	\textbf{Abstract}.
\end{center}
\vspace{.15in}

\small	Having the right assortment of shipping boxes in the fulfillment warehouse to pack and ship customer's online orders is an indispensable and integral part of nowadays eCommerce business, as it will not only help maintain a profitable business but also create  great experiences for customers. However, it is an extremely challenging operations task to strategically select the best combination of tens of box sizes from thousands of feasible ones to be responsible for hundreds of thousands of orders daily placed on millions of inventory products. In this paper, we present a machine learning approach to tackle the task by formulating the box design problem prescriptively as a generalized version of weighted $k$-medoids clustering problem, where the parameters are estimated through a variety of descriptive analytics. We test this machine learning approach on fulfillment data collected from Walmart U.S. eCommerce, and our approach is shown to be capable of improving the box utilization rate by more than $10\%$.

\vspace{.5in}
\normalsize\textbf{Keywords:} Shipping box design, $k$-medoids clustering, eCommerce, packaging science, operations research

\section{Introduction}
The assortment of shipping boxes utilized by the fulfillment warehouse to pack and ship customer's online orders is a critical component of nowadays eCommerce business, as it will directly affect not only profit margins but also customer's experience. Conventionally, many eCommerce players (e.g., walmart.com, samsclub.com and jet.com) rely on  experts' knowledge of inventory products' dimensions, levels of demand, and economic box sizes to design their assortments of shipping boxes. However, it is an extremely challenging operational task to manually select the best combination of $15$-$30$ box sizes from thousands of feasible ones for hundreds of thousands of orders daily placed on millions of products in a strategical and scalable manner. In this paper, we will propose a novel machine approach to conquer this task.

\section{Methodology}
The key idea of our approach is to model each box size as a point in space and formulate the box design problem prescriptively as a generalized version of {\em weighted $k$-medoids clustering problem}  \cite{kaufman1987clustering} to recommend $k$ box sizes. 
\\\\
Specifically, given a set of all feasible box sizes $\mc B=\set{(L_i, D_i, H_i)}_{i\in[n]}$ to select from, we solve the following optimization problem to recommend a box assortment of cardinality $k$: 
\begin{flalign}\label{eqn: gkm}
\mc S^\star = \argmin_{\mc S \subseteq [n], \; |\mc S| = k} \quad f(\mc S) := \sum_{j \in [n]} \; w_j \left( \min_{i \in \mc S}\set{c_{ij}} \right).
\end{flalign}
Here $w_j$ (the weight of each point) measures the box $j$'s economic value; and $c_{ij}$ (the generalized distance between points) measures the economic cost in substituting box $j$ with box $i$.
\\\\
In the next two subsections, we will elaborate how these parameters are estimated through a variety of descriptive analytics. At the end of this section, we will discuss our approach to solving optimization problem \eqref{eqn: gkm}.


\subsection{Box economic value $w_j$}
The box economic value is modeled to reflect its relevance to inventory products' dimension information and customers' shopping behaviors. The box $j$'s weight $w_j$ is estimated descriptively by its  (discounted) effective volume contribution, the total volume of products packed using box $j$, when packing historical customer orders in the training dataset assuming all $n$ box sizes are available.
\\\\
Specifically, we define
\begin{flalign}
w_j := \frac{{EV}_j}{{(L_j D_j H_j)}^\rho},
\end{flalign}
where $\mbox{EV}_j$ denotes the effective volume contributed by box type $B_j$ , and $\rho>0$ is a tuning parameter to penalize box sizes with large volumes. 
\\\\
In practice, we often see the number of candidate boxes $n$ ranging from $6,000$ to $8,000$. Therefore, to estimate $\set{EV_j}_{j \in [n]}$, we have to solve the bin packing problems in an extremely efficient and scalable manner. Driven by this, we develop and open-source an \textbf{\textsf{R}} package \textsf{gbp} \cite{yang2017gbp} which aims to optimize the number of boxes and the utilization rate subject to the $4$D (length, width, height and weight) constraints. This package solves $1$D-$4$D packing problem using a novelly designed best-fit-first strategy in a recursive manner; and is more powerful than previous  packing solutions \cite{lodi1999heuristic, martello2000three, lodi2002heuristic, lodi2004tspack, crainic2008extreme, crainic2009ts2pack, parreno2010hybrid, zhang2011space, zhu2012space, li2014genetic} by taking care of the weight constraint and handling the order split in packing. When compared with global optimal solutions generated by \textsf{Gorubi} \cite{optimization2013gurobi} on benchmark datasets, our solver runs more than $100$ times faster with less than 1\%-suboptimality sacrifice. 

\subsection{Box-Box economic substitution cost $c_{ij}$}
Besides the box economic value $w_j$, it is important to incorporate the substitution effects among boxes. A box type $B_i$ with small $w_i$ computed in the previous subsection could still be quite competitive if products packed by other types of boxes can be easily repacked using $B_i$ without too much sacrifice. Inspired by this, we model the box-box economic cost to reflect such substitution effects between different box sizes in packing orders. 
\\\\
Ideally, the box $i$ against box $j$ substitution cost $c_{ij}$ would indicate the extra cost of using box $i$ to pack orders, which would be optimally packed using box $j$ when all $n$ boxes are available. Thus, we define $c_{ij}$ based on the dimensional relations between box $i$ and box $j$:
\begin{flalign}
c_{ij} =
\begin{cases}
\displaystyle -\frac{l_j * d_j * h_j}{l_i*d_i*h_i}, &\mbox{for } i  \in \mc D_j \\
\displaystyle -\frac{l_j * d_j * h_j}{l_i*d_i*h_i} \left/ \left( \left\lceil\frac{l_j * d_j * h_j}{l_i*d_i*h_i} \right\rceil  + \alpha \right) \right., & \mbox{for } i \in \mc S_j \backslash \mc T_j \\
0, & \mbox{otherwise}
\end{cases},
\end{flalign}
where
\begin{flalign}
&\mc D_j := \set{i \in [n] \;\vert\; l_i \ge l_j, d_i \ge d_j, h_i \ge h_j} \nonumber \\
&\mc S_j := \set{i \in [n] \;\vert\; l_i \in [l_j-\delta, l_j+\delta], \;d_i \in [d_j-\delta, d_j+\delta], \; h_i \in [h_j-\delta, h_j+\delta]} \nonumber \\
&\mc T_j := \set{i \in [n] \;\vert\; l_i < l_j, d_i < d_j, h_i < h_j}, \nonumber
\end{flalign}
and $\alpha$ and $\delta$ are both tuning parameters.

\subsection{Generalized weighted $k$-medoids clustering problem}
The optimization problem \eqref{eqn: gkm} is a generalized version of the weighted $k$-medoids clustering problem, as the cost function $c: [n]\times [n] \to \reals$ is not necessarily a valid metric over $[n]$. Though solving the $k$-medoids problem is NP-hard,  problem \eqref{eqn: gkm} is equivalent to maximizing a nonnegative monotone submodular function subject to the cardinality constraint, which can be solved in a greedy manner with provable approximation guarantees \cite{nemhauser1978analysis, krause2014submodular}. Specifically, based on the celebrated result by \citet{nemhauser1978analysis}, the greedy approach shown in Algorithm \ref{alg: greey} provides a constant-factor approximation to the optimal solution of problem \eqref{eqn: gkm} in the sense that
\begin{flalign}
f(\overline{\mc S}) \le \left( 1-\frac{1}{e} \right) \cdot f(\mc S^\star).
\end{flalign}
\\\\
We develop and open-source another \textbf{\textsf{R}} package \textsf{skm} \cite{yang2017skm}, which efficiently implements both Algorithm \ref{alg: greey} and expectation maximization (EM) based approach to solve problem \eqref{eqn: gkm}. More generally, \textsf{skm} locates $k$ rows in an $m \times n$ matrix, such that the sum of each column minimal among the $k$ rows is minimized. For the case when $m = n$, weights are all equal and each cell value in the matrix is induced by a valid distance metric, the problem is reduced to the standard weighted $k$-medoids clustering problem. In our case, the greedy approach (i.e., Algorithm \ref{alg: greey}) can solve a selection of $20$ rows from a $6000 \times 6000$ matrix more than $100$ times faster than the EM one without loss of optimality. 
\\\\
\begin{algorithm}
	\caption{A greedy approach to solving problem \eqref{eqn: gkm}}
	\label{alg: greey}
	\begin{algorithmic}[1]
		\State Initialization: $\overline{\mc S} \gets \emptyset$, $\hat c_{ij} \gets c_{ij} \; \forall \; i,j \in[n]$
		\For{$l = 1,2,\ldots, k$}
		\State $i^\star \gets \arg\min_{i \in [n]\backslash \overline{S}} \; \sum_{j \in [n]} \hat{c}_{ij}$ \quad (with ties settled arbitrarily)
		\State $\overline{S} \gets \overline{S} \cup \set{i^\star}$
		\State $\hat c_{ij} \gets \min\set{\hat c_{ij}, \hat c_{i^\star j}} \; \forall \;i,j \in [n]$
		\EndFor
	\end{algorithmic}
\end{algorithm}

\section{Experiment}
In this experiment, we will investigate whether the box assortment designed by our machine learning approach could outperform the box assortment $\mc S_0$ that are currently using in Walmart eCommerce fulfillment centers in the U.S. 
\\\\
We choose the candidate pool $\mc B$ as all possible box sizes allowed in the fulfillment centers and carriers, and $k$ to be the same as the number of box sizes in $\mc S_0$, i.e., $k=|\mc S_0|$. The order dataset is collected from historical customers' order fulfilled by Walmart U.S. eCommerce. The dataset is randomly divided into three parts \cite{friedman2001elements}-- a training set (to train the models), a validation set (for model selection), and a test set (to assess the final model). 
\\\\
We first solve problem \eqref{eqn: gkm} using the training set under different choices of $\rho$, $\delta$ and $\alpha$. We choose $\rho \in \set{0.25, 0.50, 0.75, 1}$, $\delta \in \set{0, 1, 2, 3,4}$ and $\alpha \in \set{0, 1,2, 3, 4}$. That leads to in total 100 different models  (i.e., parameter settings) to choose from. We select the best model via estimating each box assortment's performance on the validation set. In Table \ref{tab: pt}, we report their performances in terms of both the number of boxes used in packing customer orders and utilization rate. We choose the parameter setting $\rho=0.5, \delta=4, \alpha=3$ (highlighted in Table \ref{tab: pt}) as it improves the most in utilization rate without sacrificing the number of boxes when compared with the packing result using $\mc S_0$. Finally, we solve problem (2.1) using both training and validation sets with $\rho=0.5, \delta=4, \alpha=3$ to get the ultimate box assortment $\mc S^\star$ as our recommendation. We assess the final model by comparing $\mc S^\star$ and the currently used box assortment $\mc S_0$ on the test set. The utilization is improved significantly by $10.19\%$ and the number of boxes is also reduced by $0.25\%$, which clearly demonstrates the efficacy of our machine learning approach to shipping box design.

	\begin{table}[htbp]
	\caption{\textbf{Parameter tuning over the validation set.}  For each parameter setting $(\rho, \delta, \alpha)$, we solve problem \eqref{eqn: gkm} to get $\overline{\mc S}$ using the order data in the training set, and then measuring the performance of $\overline{\mc S}$ upon orders in the validation set. Number of boxes used and utilization rate are reported by taking the current box assortment in fulfillment centers as benchmark.}
	\label{tab: pt}
	\centering
	\begin{adjustbox}{width=1\textwidth,center}
		\small
		\scalebox{1}{
			\begin{tabular}{C C C C C || C C C C C} 
				\toprule
				$\rho$
				&$\delta$
				&$\alpha$
				& Reduc. in \# of boxes (\%)
				& Imprv. in utili. rate  (\%)
				&$\rho$
				&$\delta$
				&$\alpha$
				& Reduc. in \# of boxes (\%)
				& Imprv. in utili. rate  (\%)
				\\
				\midrule
				\midrule
				0.25 & 0 & 0 & 0.33 & 9 & 0.75 & 0 & 0 & 0.2 & 7.39\\0.25 & 0 & 1 & 0.33 & 9 & 0.75 & 0 & 1 & 0.2 & 7.39\\0.25 & 0 & 2 & 0.33 & 9 & 0.75 & 0 & 2 & 0.2 & 7.39\\0.25 & 0 & 3 & 0.33 & 9 & 0.75 & 0 & 3 & 0.2 & 7.39\\0.25 & 0 & 4 & 0.33 & 9 & 0.75 & 0 & 4 & 0.2 & 7.39\\0.25 & 1 & 0 & 0.33 & 9.05 & 0.75 & 1 & 0 & 0.19 & 5.39\\0.25 & 1 & 1 & 0.33 & 8.92 & 0.75 & 1 & 1 & 0.2 & 6.59\\0.25 & 1 & 2 & 0.33 & 9 & 0.75 & 1 & 2 & 0.2 & 6.58\\0.25 & 1 & 3 & 0.33 & 9 & 0.75 & 1 & 3 & 0.2 & 6.58\\0.25 & 1 & 4 & 0.33 & 9 & 0.75 & 1 & 4 & 0.2 & 6.58\\0.25 & 2 & 0 & 0.31 & 7.4 & 0.75 & 2 & 0 & 0.28 & 6.11\\0.25 & 2 & 1 & 0.31 & 9.42 & 0.75 & 2 & 1 & 0.24 & 5.38\\0.25 & 2 & 2 & 0.33 & 9.11 & 0.75 & 2 & 2 & 0.28 & 5.49\\0.25 & 2 & 3 & 0.33 & 9.11 & 0.75 & 2 & 3 & 0.24 & 6.66\\0.25 & 2 & 4 & 0.33 & 9.11 & 0.75 & 2 & 4 & 0.28 & 5.22\\0.25 & 3 & 0 & 0.11 & 7.93 & 0.75 & 3 & 0 & 0.1 & 4.74\\0.25 & 3 & 1 & 0.31 & 9.35 & 0.75 & 3 & 1 & 0.2 & 6.31\\0.25 & 3 & 2 & 0.33 & 9.05 & 0.75 & 3 & 2 & 0.2 & 6.35\\0.25 & 3 & 3 & 0.33 & 8.94 & 0.75 & 3 & 3 & 0.2 & 7.35\\0.25 & 3 & 4 & 0.32 & 9.13 & 0.75 & 3 & 4 & 0.2 & 7.35\\0.25 & 4 & 0 & 0.02 & 5.02 & 0.75 & 4 & 0 & 0.02 & 2.98\\0.25 & 4 & 1 & 0.3 & 9.3 & 0.75 & 4 & 1 & 0.2 & 6.09\\0.25 & 4 & 2 & 0.3 & 9.67 & 0.75 & 4 & 2 & 0.2 & 7.59\\0.25 & 4 & 3 & 0.33 & 8.54 & 0.75 & 4 & 3 & 0.2 & 7.56\\0.25 & 4 & 4 & 0.33 & 8.73 & 0.75 & 4 & 4 & 0.2 & 7.71\\
				\hline
				\hline
				0.5 & 0 & 0 & 0.28 & 8.43 & 1 & 0 & 0 & 0.2 & 3.99\\0.5 & 0 & 1 & 0.28 & 8.43 & 1 & 0 & 1 & 0.2 & 3.99\\0.5 & 0 & 2 & 0.28 & 8.43 & 1 & 0 & 2 & 0.2 & 3.99\\0.5 & 0 & 3 & 0.28 & 8.43 & 1 & 0 & 3 & 0.2 & 3.99\\0.5 & 0 & 4 & 0.28 & 8.43 & 1 & 0 & 4 & 0.2 & 3.99\\0.5 & 1 & 0 & 0.29 & 8.36 & 1 & 1 & 0 & 0.18 & 4.25\\0.5 & 1 & 1 & 0.28 & 8.87 & 1 & 1 & 1 & 0.2 & 3.58\\0.5 & 1 & 2 & 0.28 & 8.47 & 1 & 1 & 2 & 0.2 & 3.99\\0.5 & 1 & 3 & 0.28 & 8.47 & 1 & 1 & 3 & 0.2 & 3.99\\0.5 & 1 & 4 & 0.28 & 8.47 & 1 & 1 & 4 & 0.2 & 3.99\\0.5 & 2 & 0 & 0.29 & 6.55 & 1 & 2 & 0 & 0.12 & 2.17\\0.5 & 2 & 1 & 0.29 & 8.97 & 1 & 2 & 1 & 0.2 & 3.3\\0.5 & 2 & 2 & 0.3 & 8.8 & 1 & 2 & 2 & 0.2 & 4.19\\0.5 & 2 & 3 & 0.28 & 8.47 & 1 & 2 & 3 & 0.2 & 2.76\\0.5 & 2 & 4 & 0.28 & 8.47 & 1 & 2 & 4 & 0.2 & 2.09\\0.5 & 3 & 0 & 0.13 & 8.6 & 1 & 3 & 0 & 0.22 & 1.82\\0.5 & 3 & 1 & 0.29 & 8.24 & 1 & 3 & 1 & 0.03 & 2.28\\0.5 & 3 & 2 & 0.29 & 8.7 & 1 & 3 & 2 & 0.03 & 3.05\\0.5 & 3 & 3 & 0.29 & 9.09 & 1 & 3 & 3 & 0.28 & 2.89\\0.5 & 3 & 4 & 0.3 & 9.66 & 1 & 3 & 4 & 0.28 & 2.81\\0.5 & 4 & 0 & 0.02 & 4.33 & 1 & 4 & 0 & -0.34 & 1.42\\0.5 & 4 & 1 & 0.21 & 8.33 & 1 & 4 & 1 & 0.18 & 0.75\\0.5 & 4 & 2 & 0.2 & 9.26 & 1 & 4 & 2 & 0.2 & 3.92\\
				\textbf{0.5} & 	\textbf{4} & \textbf{3} & \textbf{0.2} & \textbf{10.28} & 1 & 4 & 3 & 0.02 & 2.28\\0.5 & 4 & 4 & 0.28 & 8.89 & 1 & 4 & 4 & 0.03 & 3.07\\
				\bottomrule
			\end{tabular}
		}
	\end{adjustbox}
\end{table}

\newpage
\section{Future Work}
In this paper, we formulate the shipping box design problem as a generalized version of weighted $k$-medoids clustering problem, of which the parameters are analytically estimated through customers' historical order data.  In the future, we plan to include more variations in defining the weight function $\set{w_i}_{i\in [n]}$ and the substitution cost function $\set{c_{ij}}_{i, j\in[n]}$ to take full advantage of  our machine learning approach in designing shipping boxes.

\section*{Acknowledgment}
We are grateful to Iris Zhang, Aliasgar Kutiyanawala, John Yan and Nate Faust for helpful discussions in both packing algorithms and business insights, and to Anran Li, who brought the paper \cite{nemhauser1978analysis} to our attention. We are also grateful to the anonymous reviewers for their helpful suggestions and comments that substantially improve the paper.



\bibliographystyle{alpha}
\bibliography{bxd}

\end{document}